\documentclass[runningheads]{llncs}
\usepackage{graphicx}
\usepackage{amsmath,amssymb} 
\usepackage{color}
\usepackage{subcaption}
\usepackage{algorithm}
\usepackage{algpseudocode}
\usepackage[width=122mm,left=12mm,paperwidth=146mm,height=193mm,top=12mm,paperheight=217mm]{geometry}
\begin{document}

\title{Estimating Depth from RGB and Sparse Sensing}
\titlerunning{Estimating Depth from RGB and Sparse Sensing}

\author{Zhao Chen \and Vijay Badrinarayanan \and Gilad Drozdov \and Andrew Rabinovich}
\authorrunning{Z. Chen, V. Badrinarayanan, G. Drozdov, and A. Rabinovich}
\institute{Magic Leap, Sunnyvale CA 94089, USA\\
\email{\{zchen, vbadrinarayanan, gdrozdov, arabinovich\}@magicleap.com}}

\maketitle

\begin{abstract}

We present a deep model that can accurately produce dense depth maps given an RGB image with known depth at a very sparse set of pixels. The model works \textit{simultaneously} for both indoor/outdoor scenes and produces state-of-the-art dense depth maps at nearly real-time speeds on both the NYUv2 and KITTI datasets. We surpass the state-of-the-art for monocular depth estimation even with depth values for only 1 out of every $\sim 10000$ image pixels, and we outperform other sparse-to-dense depth methods at all sparsity levels. With depth values for $1/256$ of the image pixels, we achieve a mean error of less than $1\%$ of actual depth on indoor scenes, comparable to the performance of consumer-grade depth sensor hardware. Our experiments demonstrate that it would indeed be possible to efficiently transform sparse depth measurements obtained using e.g. lower-power depth sensors or SLAM systems into high-quality dense depth maps.

\end{abstract}

\section{Introduction}
Efficient, accurate and real-time depth estimation is essential for a wide variety of scene understanding applications in domains such as virtual/mixed reality, autonomous vehicles, and robotics. Currently, a consumer-grade Kinect v2 depth sensor consumes $\sim$ 15W of power, only works indoors at a limited range of $\sim 4.5m$, and degrades under increased ambient light \cite{horaud2016overview}. For reference, a future VR/MR head mounted depth camera would need to consume $1/100$th the power and have a range of 1-80m (indoors and outdoors) at the full FOV and resolution of an RGB camera. Such requirements present an opportunity to jointly develop energy-efficient depth hardware and depth estimation models. Our work begins to address depth estimation from this perspective.

Due to its intrinsic scale ambiguity, monocular depth estimation is a challenging problem, with state-of-the-art models \cite{eigen2015predicting,laina2016deeper} still producing $>12\%$ mean absolute relative error on the popular large-scale NYUv2 indoor dataset \cite{Silberman:ECCV12}. Such errors are prohibitive for applications such as 3D reconstruction or tracking, and fall very short of depth sensors such as the Kinect that boast relative depth error on the order of $\sim 1\%$ \cite{khoshelham2012accuracy,nguyen2012modeling} indoors.

\begin{figure}[t]
\centering
\includegraphics[scale=0.28]{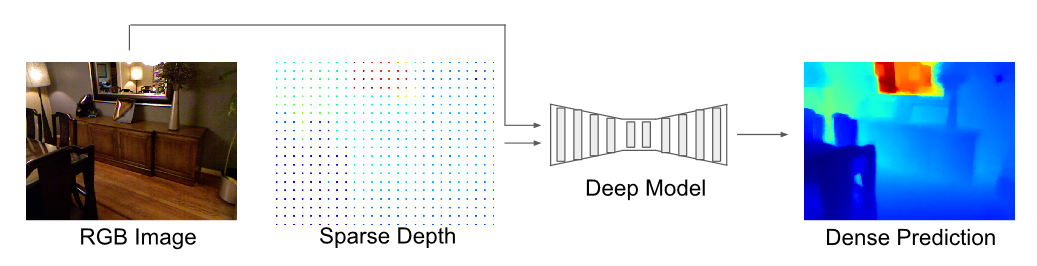}
\caption{\textbf{From Sparse to Dense Depth.} An RGB Image and very sparse depth map are input into a deep neural network. We obtain a high-quality dense depth prediction as our final output.}
\label{fig:problem_setup}
\end{figure}

Acknowledging the limitations of monocular depth estimation, we provide our depth model with a sparse amount of measured depth along with an RGB image (See Fig. \ref{fig:problem_setup}) in order to estimate the full depth map. Such sparse depth resolves the depth scale ambiguity, and could be obtained from e.g. a sparser illumination pattern in Time-of-Flight sensors \cite{horaud2016overview}, confident stereo matches, LiDAR-like sensors, or a custom-designed sparse sensor. We show that the resultant model can provide comparable performance to a modern depth sensor, despite only observing a small fraction of the depth map. We believe our results can thus motivate the design of smaller and more energy-efficient depth sensor hardware. As the objective is now to \textit{densify} a sparse depth map (with additional cues from an RGB image), we call our model Deep Depth Densification, or D$^3$. 

One advantage of our D$^3$ model is that it accommodates for arbitrary sparse depth input patterns, each of which may correspond to a relevant physical system. A regular grid of sparse depth may come from a lower-power depth sensor, while certain interest point sparse patterns such as ORB \cite{rublee2011orb} or SIFT \cite{lowe2004distinctive} could be output from modern SLAM systems \cite{ma2017sparse}. In the main body of this work, we will focus on regular grid patterns due to their ease of interpretation and immediate relevance to existing depth sensor hardware, although we detail experiments on ORB sparse patterns in the Supplementary Materials. 

Our contributions to the field of depth estimation are as follows: 
\begin{enumerate}
\item A deep network model for dense scene depth estimation that achieves accuracies comparable to conventional depth sensors. 
\item A depth estimation model which works \textit{simultaneously} for indoors and outdoors scenes and is robust to common measurement errors.
\item A flexible, invertible method of parameterizing sparse depth inputs that can accommodate arbitrary sparse input patterns during training and testing.

\end{enumerate}


\section{Related Work}

Depth estimation has been tackled in computer vision well before the advent of deep learning \cite{saxena2006learning,sinz2004learning}; however, the popularization of encoder-decoder deep net architectures \cite{badrinarayanan2017segnet,long2015fully}, which produce full-resolution pixel-wise prediction maps, make deep neural networks particularly well-suited for this task. Such advances have spurred a flurry of research into deep methods for depth estimation, whether through fusing CRFs with deep nets \cite{xu2017multi}, leveraging geometry and stereo consistency \cite{garg2016unsupervised,kuznietsov2017semi}, or exploring novel deep architectures \cite{laina2016deeper}. 

Depth in computer vision is often used as a component for performing other perception tasks. One of the first approaches to deep depth estimation also simultaneously estimates surface normals and segmentation in a multitask architecture \cite{eigen2015predicting}. Other multitask vision networks \cite{chen2017gradnorm,kendallarxiv,teichmann2016multinet} also commonly use depth as a complementary output to benefit overall network performance. Using depth as an explicit input is also common in computer vision, with plentiful applications in tracking \cite{song2013tracking,teichman_tracking}, SLAM systems \cite{slam2,slam1} and 3d reconstruction/detection \cite{hermans2014dense,lin2013holistic}. There is clearly a pressing demand for high-quality depth maps, but current depth hardware solutions are power-hungry, have severe range limitations \cite{horaud2016overview}, and the current traditional depth estimation methods \cite{eigen2015predicting,laina2016deeper} fail to achieve the accuracies necessary to supersede such hardware.

Such challenges naturally lead to depth densification, a middle ground that combines the power of deep learning with energy-efficient sparse depth sensors. Depth densification is related to depth superresolution \cite{hui2016depth,song2016deep}, but superresolution generally uses a bilinear or bicubic downsampled depth map as input, and thus still implicitly contains information from all pixels in the low-resolution map. This additional information would not be accessible to a true sparse sensor, and tends to make the estimation problem easier (see Supplementary Material). Work in \cite{lu2015sparse} and \cite{ma2017sparse} follows the more difficult densification paradigm where only a few pixels of measured depth are provided. We will show that our densification network outperforms the methods in both \cite{lu2015sparse} and \cite{ma2017sparse}.

\section{Methodology}

\subsection{Input Parametrization for Sparse Depth Inputs}\label{sec:parametrization}
We desire a parametrization of the sparse depth input that can accommodate arbitrary sparse input patterns. This should allow for varying such patterns not only across different deep models but even within the same model during training and testing. Therefore, rather than directly feeding a highly discontinuous sparse depth map into our deep depth densification (D$^3$) model (as in Fig. \ref{fig:problem_setup}), we propose a more flexible parametrization of the sparse depth inputs. 

At each training step, the inputs to our parametrization are: 

\begin{enumerate}
\item $I(x,y)$ and $D(x,y)$: RGB vector-valued image $I$ and ground truth depth $D$. Both maps have dimensions H$\times$W. Invalid values in $D$ are encoded as zero.
\item $M(x,y)$: Binary pattern mask of dimensions H$\times$W, where $M(x,y)=1$ defines $(x,y)$ locations of our desired depth samples. $M(x,y)$ is preprocessed so that all points where $M(x,y)=1$ must correspond to valid depth points ($D(x,y)>0$). (see Algorithm \ref{alg:sparse_input}). 
\end{enumerate}

\begin{figure}[t!]
\centering
\includegraphics[scale=0.395]{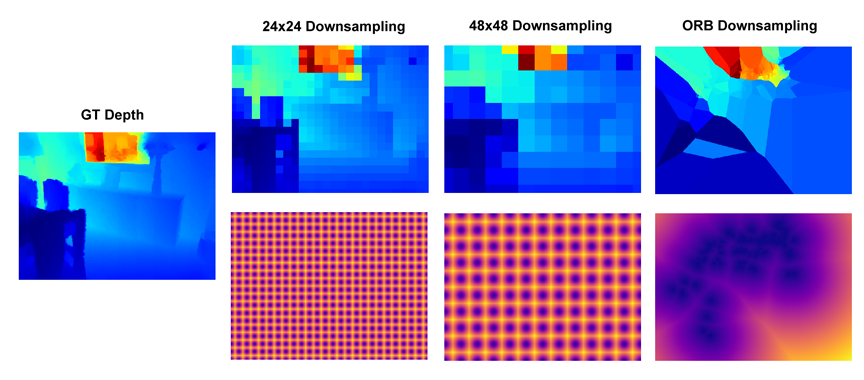}
\caption{\textbf{Various Sparse Patterns.} NN fill maps $\mathcal{S}_1$ (top row) and the sampling pattern Euclidean Distance transforms $\mathcal{S}_2$ (bottom row) are shown for both regular and irregular sparse patterns. Dark points in $\mathcal{S}_2$ correspond to the pixels where we have access to depth information.}
\label{fig:sampling_patterns}
\end{figure}

\noindent From $I$, $D$, and $M$, we form \textit{two maps} for the sparse depth input, $\mathcal{S}_1(x,y)$ and $\mathcal{S}_2(x,y)$. Both maps have dimension H$\times$W (see Fig. \ref{fig:sampling_patterns} for examples). 
\begin{itemize}
\item$\mathcal{S}_1(x,y)$ is a NN (nearest neighbor) fill of the sparse depth $M(x,y)*D(x,y)$. 

\item$\mathcal{S}_2(x,y)$ is the Euclidean Distance Transform of $M(x,y)$, i.e. the L$_2$ distance between (x,y) and the closest point (x',y') where $M(x',y')=1$. 
\end{itemize}

The final parametrization of the sparse depth input is the concatenation of $\mathcal{S}_1(x,y)$ and $\mathcal{S}_2(x,y)$, with total dimension H$\times$W$\times$2. This process is described in Algorithm $\ref{alg:sparse_input}$. The parametrization is fast and involves at most two Euclidean Transforms. The resultant NN map $\mathcal{S}_1$ is nonzero everywhere, allowing us to treat the densification problem as a \textit{residual prediction} with respect to $\mathcal{S}_1$. The distance map $\mathcal{S}_2$ informs the model about the pattern mask $M(x,y)$ and acts as a prior on the residual magnitudes the model should output (i.e. points farther from a pixel with known depth tend to incur higher residuals). Inclusion of $\mathcal{S}_2$ can substantially improve model performance and training stability, especially when multiple sparse patterns are used during training (see Section \ref{sec:multires}).

\begin{algorithm}[t!]
   \caption{Sparse Inputs for the Deep Depth Densification (D$^3$) Model}
   \label{alg:sparse_input}
\begin{algorithmic}
\item INPUT image $I(x,y)$, depth $D(x,y)$, and pattern mask $M(x,y)$. 
\item INITIALIZE $\mathcal{S}_1(x,y)=0$, $\mathcal{S}_2(x,y)=0$ for all $(x,y)$.
\item FOR $\mathbf{r}:=(x,y)$ s.t. $D(\bold{r})=0$ AND $M({\mathbf{r}})=1$:
\begin{algorithmic}
\item $\bold{r}_{new} = \text{argmin}_{\bold{r}'}||\bold{r}-\bold{r}'||_2 \text{ s.t. } D(\bold{r}')>0$; \hfill(\textit{$||\cdot||_2$ denotes the $L_2$ norm.})
\item $M(\bold{r}) = 0$;\indent  $M(\bold{r}_{new}) = 1$;
\end{algorithmic}
\item ENDFOR \hfill(\textit{All depth locations are now valid.})
\item FOR $\bold{r}:=(x,y)$:
\begin{algorithmic}
\item $\bold{r}_{\text{nearest}} = \text{argmin}_{\bold{r}'}||\bold{r}'-\bold{r}||_2 \text{ s.t. } M(\bold{r}')=1$;
\item $\mathcal{S}_1(x,y) = D(\bold{r}_{nearest})$; \indent $\mathcal{S}_2(x,y) = \sqrt{||\bold{r}_{nearest}-\bold{r}||_2}$;
\end{algorithmic}
\item ENDFOR
\item OUTPUT \textbf{concatenate}$(\mathcal{S}_1,\mathcal{S}_2)$
\end{algorithmic}
\end{algorithm}

In this work, we primarily focus on regular grid patterns, as they are high-coverage sparse maps that enable straightforward comparisons to prior work (as in \cite{lu2015sparse}) which often assume a grid-like sparse pattern, but our methods fully generalize to other patterns like ORB (see Supplementary Materials).

\subsection{Sparse Pattern Selection}\label{sec:sampling_strategies}

For regular grid patterns, we try to ensure minimal spatial bias when choosing the pattern mask $M(x,y)$ by enforcing equal spacing between subsequent pattern points in both the $x$ and $y$ directions. This results in a checkerboard pattern of square regions in the sparse depth map $\mathcal{S}_1$ (see Fig. \ref{fig:sampling_patterns}). Such a strategy is convenient when one deep model must accommodate images of different resolutions, as we can simply extend the square pattern in $M(x,y)$ from one resolution to the next. For ease of interpretation, we will always use sparse patterns close to an integer level of downsampling; for a downsampling factor of $A\times A$, we take $\sim H*W/A^2$ depth values as the sparse input. For example, for 24$\times$24 downsampling on a 480$\times$640 image, this would be 0.18$\%$ of the total pixels.  

Empirically we observed that it is beneficial to vary the sparse pattern $M(x,y)$ during training. For a desired final pattern of $N$ sparse points, we employ a slow decay learning schedule following $N_{\text{sparse}}(t) = \lfloor 5Ne^{-0.0003t} + N\rfloor$ for training step $0\leq t\leq 80000$. Such a schedule begins training at six times the desired sparse pattern density and smoothly decays towards the final density as training progresses. Compared to a static sparse pattern, we see a relative decrease of $\sim 3\%$ in the training L$_2$ loss and also in the mean relative error when using this decay schedule. We can also train with randomly varying sampling densities at each training step. This we show in Section \ref{sec:multires} results in a deep model which performs well \textit{simultaneously} at different sampling densities.

\section{Experimental Setup}
\begin{figure}[htb!]
\centering
\includegraphics[scale=0.21]{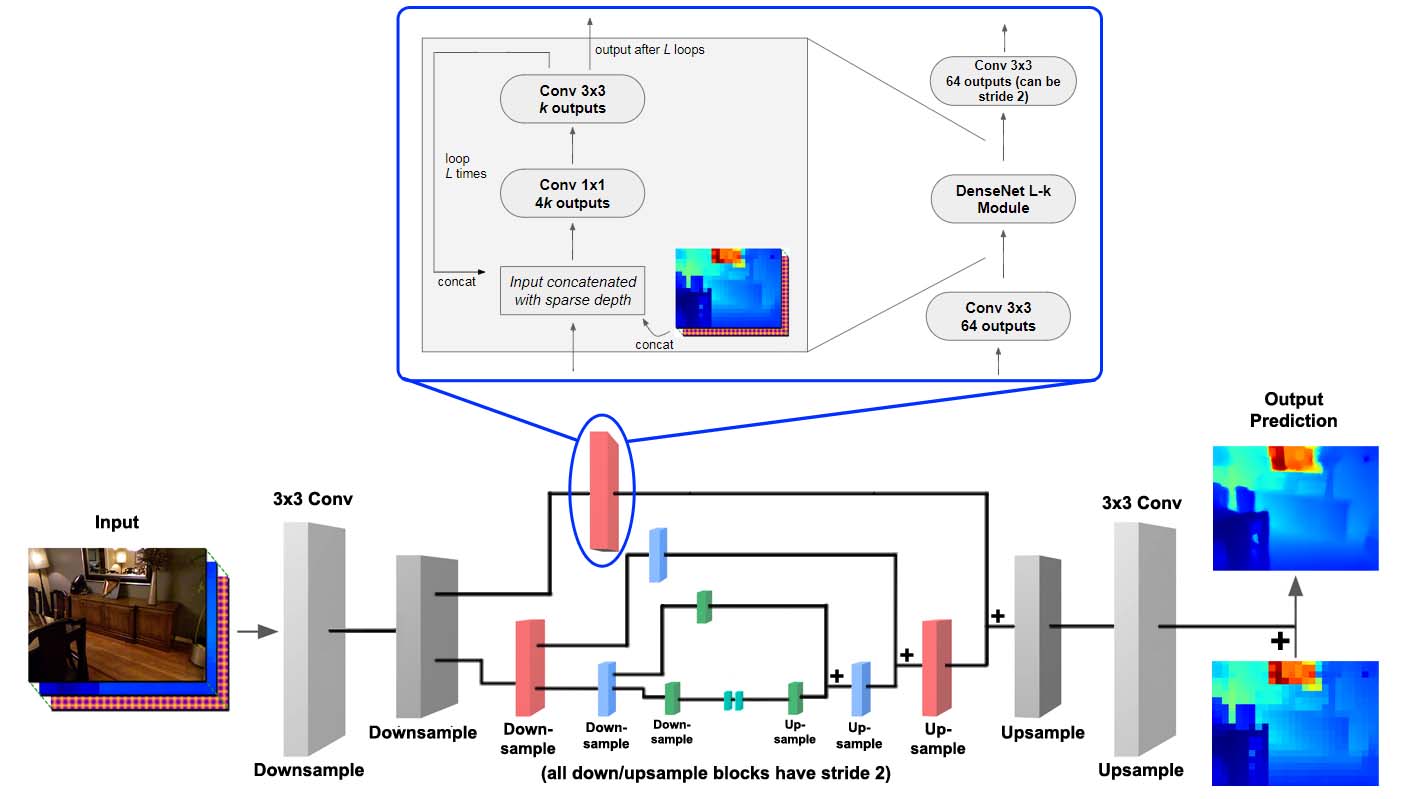}
\caption{\textbf{D$^3$ Network Architecture.} Our proposed multi-scale deep network takes an RGB image concatenated with $\mathcal{S}_1$ and $\mathcal{S}_2$ as inputs. The first and last computational blocks are simple 3x3 stride-2 convolutions, but all other blocks are DenseNet modules \cite{huang2017densely} (see inset). All convolutional layers in the network are batch normalized \cite{ioffe2015batch} and ReLU activated. The network outputs a residual that is added to the sparse depth map $\mathcal{S}_1$ to produce the final dense depth prediction.}
\label{fig:arch}
\end{figure}

\subsection{Architecture}

We base our network architecture (see Fig. \ref{fig:arch}) on the network used in \cite{chen2016single} but with DenseNet \cite{huang2017densely} blocks in place of Inception \cite{szegedy2015going} blocks. We empirically found it critical for our proposed model to carry the sparse depth information throughout the deep network, and the residual nature of DenseNet is well-suited for this requirement. For optimal results, our architecture retains feature maps at multiple resolutions for addition back into the network during the decoding phase. 

Each block in Fig. \ref{fig:arch} represents a DenseNet Module (see Fig. \ref{fig:arch} inset for a precise module schematic) except for the first and last blocks, which are simple 3x3 stride-2 convolutional layers. A copy of the sparse input $[\mathcal{S}_1,\mathcal{S}_2]$ is presented as an additional input to each module, downsampled to the appropriate resolution. Each DenseNet module consists of $2L$ layers and $k$ feature maps per layer; we use $L=5$ and $k=12$. At downsample/upsample blocks, the final convolution has stride 2.  The (residual) output of the network is added to the sparse input map $\mathcal{S}_1$ to obtain the final depth map estimate. 

\subsection{Datasets}

We experiment extensively with both indoor and outdoor scenes. For indoor scenes, we use the NYUv2 \cite{Silberman:ECCV12} dataset, which provides high-quality 480$\times$640 depth data taken with a Kinect V1 sensor with a range of up to 10m. Missing depth values are filled using a standard approach \cite{levin2004colorization}. We use the official split of 249/215 train/validation scenes, and sample 26331 images from the training scenes. We further augment the training set with horizontal flips. We test on the standard validation set of 654 images to compare against other methods. 

For outdoor scenes, we use the KITTI road scenes dataset \cite{Uhrig2017THREEDV}, which has a depth range up to $\sim$85m. KITTI provides over 80000 images for training, which we further augment with horizontal flips. We test on the full validation set ($\sim$10\% the size of the training set). KITTI images have resolution 1392$\times$512, but we take random 480$\times$640 crops during training to enable joint training with NYUv2 data. The 640 horizontal pixels are sampled randomly while the 480 vertical pixels are the 480 bottom pixels of the image (as KITTI only provides LiDAR GT depth towards ground level). The LiDAR projections used in KITTI result in very sparse depth maps (with only $\sim$10\% of depths labeled per image), and we only evaluate our models on points with GT depth. 

\subsection{General Training Characteristics and Performance Metrics}
In all our experiments we train with a batch size of 8 across 4 Maxwell Titan X GTX GPUs using Tensorflow 1.2.1. We train for 80000 batches and start with a learning rate of 1e-3, decaying the learning rate by 0.2 every 25000 steps. We use Adam \cite{kingma2014adam} as our optimizer, and standard pixel-wise $L_2$ loss to train.

Standard metrics are used \cite{eigen2015predicting,ma2017sparse} to evaluate our depth estimation model against valid GT depth values. Let $\hat{y}$ be the predicted depth and $y$ the GT depth for $N$ pixels in the dataset. We measure: (1) Root Mean Square Error (RMSE): $\sqrt{\frac{1}{N}\sum [\hat{y}-y]^2} $, (2) Mean Absolute Relative Error (MRE): $\frac{100}{N}\sum \left(\frac{|\hat{y}-y|}{y}\right)$, and (3) Delta Thresholds ($\delta_i$): $\frac{|\{\hat{y}|\text{max}(\frac{y}{\hat{y}},\frac{\hat{y}}{y})<1.25^i\}|}{|\{\hat{y}\}|}$. $\delta_i$ is the percentage of pixels with relative error under a threshold controlled by the constant $i$.

\section{Results and Analysis}
Here we present results and analysis of the D$^3$ model for both indoor (NYUv2) and outdoor (KITTI) datasets. We further demonstrate that D$^3$ is robust to input errors and also generalizes to multiple sparse input patterns.

\subsection{Indoor scenes from NYUv2}\label{sec:NYU_results}

\begin{table}[htb!]
\caption{\textbf{D$^3$ Performance on NYUv2}. Lower RMSE and MRE is better, while higher $\delta_i$ is better. NN Fill corresponds to using the sparse map $\mathcal{S}_1$ as our final prediction. If no sparse depth is provided, the D$^3$ model falls short of \cite{eigen2015predicting} and \cite{laina2016deeper}, but even at 0.01\% points sampled the D$^3$ model offers significant improvements over state-of-the-art non-sparse methods. D$^3$ additionally performs the best compared to other sparse depth methods at all input sparsities.}
\begin{center}
\label{table:NYU_results}
\begin{tabular}{llllllll}
\hline\noalign{\smallskip}
Model   & \% Points & Downsampling&RMSE  & MRE  & $\delta_1$ & $\delta_2$ & $\delta_3$ \\
& Sampled & Factor &(m)&(\%)&(\%)&(\%)&(\%)\\
\noalign{\smallskip}
\hline
\noalign{\smallskip}
Eigen et al. \cite{eigen2015predicting} & 0& N/A &  0.641&15.8&76.9&95.0&98.8 \\
Laina et al. \cite{laina2016deeper} & 0& N/A & \textbf{0.573} & \textbf{12.7} & \textbf{81.1}& \textbf{95.3} & \textbf{98.8} \\
D$^3$ No Sparse & 0& N/A &  0.711&22.37&67.32&89.68&96.73 \\
\noalign{\smallskip}
\hline
\noalign{\smallskip}
NN Fill & 0.011& 96$\times$96 &  0.586&11.69&86.8&95.8&98.4 \\
D$^3$ (Ours) & 0.011& 96$\times$96 &  \textbf{0.318}&\textbf{7.20}&\textbf{94.2}&\textbf{98.9}&\textbf{99.8} \\
\noalign{\smallskip}
\hline
\noalign{\smallskip}
Ma et al. \cite{ma2017sparse} & 0.029 & $\sim$59$\times$59 & 0.351 & 7.8 & 92.8 & 98.4 & 99.6\\
NN Fill & 0.043& 48$\times$48 &  0.383&6.23&94.42&98.20&99.35\\
D$^3$ Mixed (Ours) & 0.043& 48$\times$48 &  0.217&3.77&97.90&99.65&99.93\\
D$^3$ (Ours) & 0.043& 48$\times$48 &  \textbf{0.193}&\textbf{3.21}&\textbf{98.31}&\textbf{99.73}&\textbf{99.95}\\
\noalign{\smallskip}
\hline
\noalign{\smallskip}
NN Fill & 0.174& 24$\times$24 & 0.250 & 3.20 & 97.5 & 99.3 & 99.8 \\
Lu et al. \cite{lu2015sparse} & -& 24$\times$24 & 0.171 & - & - & - & - \\
D$^3$ Mixed (Ours) & 0.174& 24$\times$24 & 0.131 & 1.76 & 99.31 & 99.90 & 99.98 \\
D$^3$ (Ours) & 0.174& 24$\times$24 & \textbf{0.118} & \textbf{1.49} & \textbf{99.45} & \textbf{99.92} & \textbf{99.98} \\
\noalign{\smallskip}
\hline
\noalign{\smallskip}
Ma et al. \cite{ma2017sparse} & 0.289 & $\sim$19$\times$19 & 0.23 & 4.4 & 97.1 & 99.4 & 99.8\\
NN Fill & 0.391& 16$\times$16 &  0.192& 2.10 & 98.5 & 99.6 & 99.88 \\
Lu et al. \cite{lu2015sparse} & -& 16$\times$16 & 0.108 & - & - & - & - \\
D$^3$ (Ours) & 0.391& 16$\times$16 &  \textbf{0.087}&\textbf{0.99}&\textbf{99.72}&\textbf{99.97}&\textbf{99.99} \\
\hline 
\end{tabular}
\end{center}
\end{table}

\begin{figure}[htb!]
\centering
	
	\begin{subfigure}[h]{0.49\linewidth}
		\includegraphics[width=\textwidth]{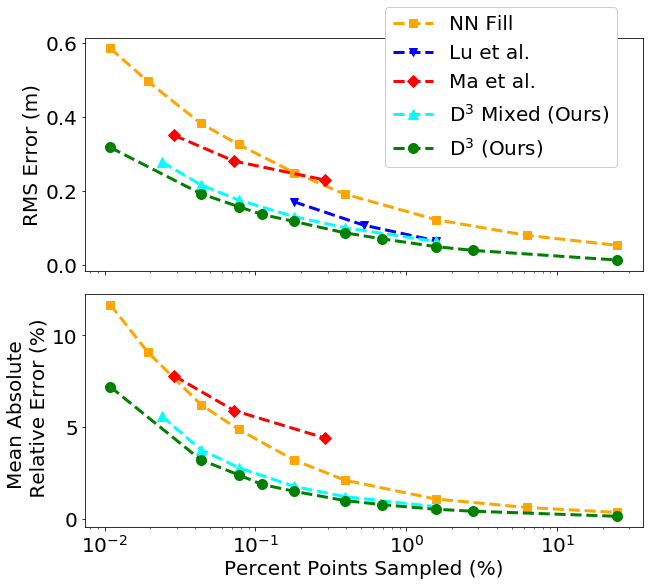}       						   
	\end{subfigure}
	~
	\begin{subfigure}[h]{0.39\linewidth}
		\includegraphics[width=\textwidth]{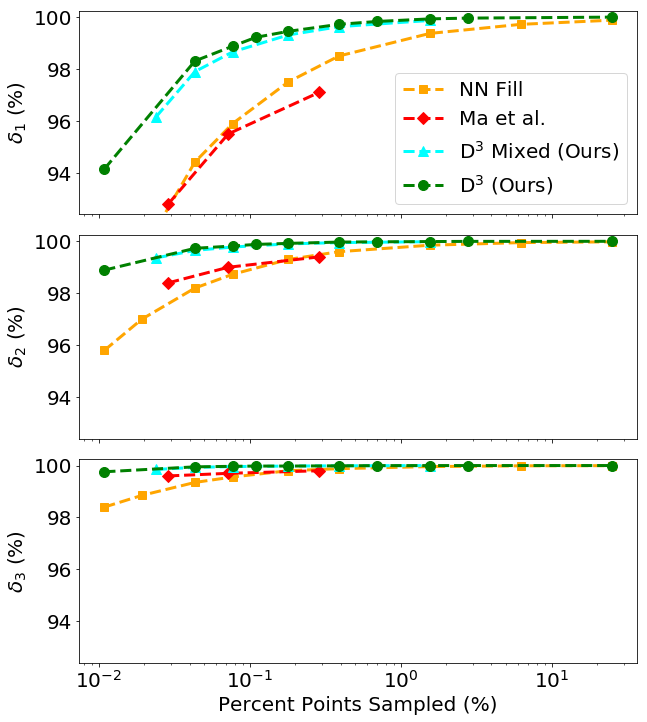}
	\end{subfigure}
	\caption{\textbf{Performance on the NYUv2 Dataset.} RMSE and MRE are plotted on the left (lower is better), while the $\delta_i$ are plotted on the right (higher is better). Our D$^3$ models achieve the best performance at all sparsities, while joint training on outdoor data (D$^3$ mixed) only incurs a minor performance loss.}
	\label{fig:nyu_plots}
\end{figure}

From Table \ref{table:NYU_results} we see that, at all pattern sparsities, the D$^3$ network offers superior performance for all metrics\footnote{A model trained with 0.18\% sparsity performs very well on a larger NYUv2 test set of 37K images: RMSE 0.116m/ MRE 1.34\%/$\delta_1$ 99.52\%/$\delta_2$ 99.93\%/$\delta_3$ 99.986\%.} compared to the results in \cite{ma2017sparse} and in \cite{lu2015sparse}\footnote{As results in \cite{lu2015sparse} were computed on a small subset of the NYUv2 val set, metrics were normalized to each work's reported NN fill RMSE to ensure fair comparison.}. The accuracy metrics for the D$^3$ \textit{mixed} network represent the NYUv2 results for a network that has been simultaneously trained on the NYUv2 (indoors) and KITTI (outdoors) datasets (more details in Section \ref{sec:KITTI_results}). We see that incorporating an outdoors dataset with significantly different semantics only incurs a mild degradation in accuracy. Fig. \ref{fig:nyu_plots} has comparative results for additional sparsities, and once again demonstrates that our trained models are more accurate than other recent approaches.

\begin{figure}[H]
	\centering
	\begin{subfigure}[h]{0.47\linewidth}
		\includegraphics[width=\textwidth]{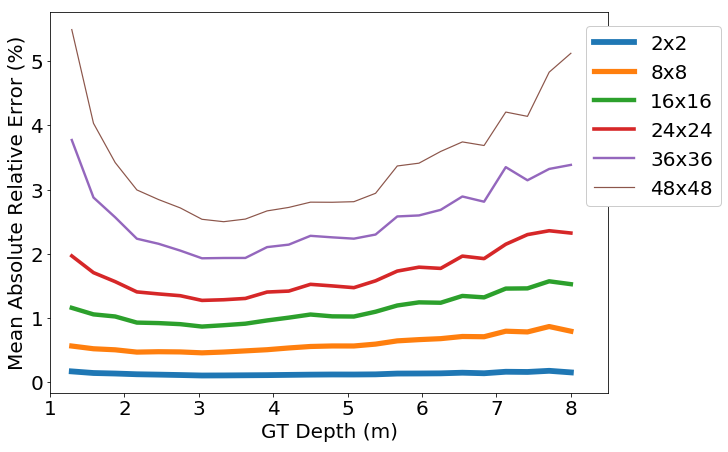}   
        \caption{}
	\end{subfigure}
	~
	\begin{subfigure}[h]{0.4\linewidth}
		\includegraphics[width=\textwidth]{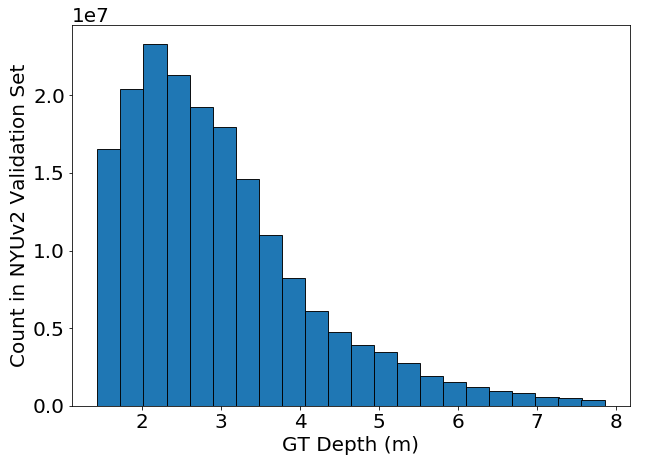}
        \caption{}
	\end{subfigure}
	\caption{\textbf{NYUv2 MRE performance at different depths.} (a) MRE at different depths for varying levels of sparsity. At 0.39\% sparsity the average MRE is less than 1\% which is comparable to depth sensors. (b) Histogram of GT depths in the validation dataset; higher relative errors correspond to rarer depth values.}
	\label{fig:nyu_mre}
\end{figure}

At 16$\times$16 downsampling our absolute mean relative error falls below 1$\%$ (at 0.99\%). At this point, the error of our D$^3$ model becomes comparable to the error in consumer-grade depth sensors \cite{horaud2016overview}. Fig. \ref{fig:nyu_mre}(a) presents a more detailed plot of relative error at different values of GT depth. Our model performs well at most common indoor depths (around 2-4m), as can be assessed from the histogram in Fig. \ref{fig:nyu_mre}(b). At farther depths the MRE deteriorates, but these depth values are rarer in the dataset. This suggests that using a more balanced dataset can improve those MRE values as well.
\newpage


\begin{figure}[htb!]
\centering
\includegraphics[scale=0.39]{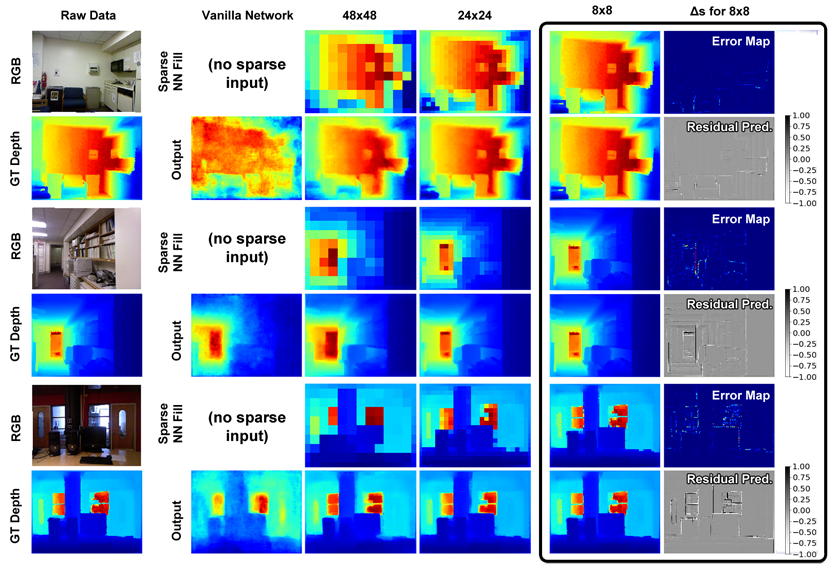}
\caption{\textbf{Visualization of D$^3$ Predictions on NYUv2.} Left column: Sample RGB and GT depths. Middle columns: sparse $\mathcal{S}_1$ map on top and D$^3$ network prediction on bottom for different sparsities. Vanilla network denotes the case with no sparse input (monocular depth estimation). Final column: D$^3$ residual predictions (summed with $\mathcal{S}_1$ to obtain the final prediction) and error maps of the final estimate with respect to GT. Errors are larger at farther distances. Residuals are plotted in grayscale and capped at $|\delta|\leq 1$ for better visualization; they exhibit similar sharp features as $\mathcal{S}_1$, showing how a D$^3$ model cancels out the nonsmoothness of $\mathcal{S}_1$.}
\label{fig:NYU_vis}
\end{figure}

Visualizations of our network predictions on the NYUv2 dataset are shown in Fig. \ref{fig:NYU_vis}. At a highly sparse 48$\times$48 downsampling, our D$^3$ network already shows a dramatic improvement over a vanilla network without any sparse input. We note here that although network outputs are added as residuals to a sparse map with many first order discontinuities, the final predictions appear smooth and relatively free of sharp edge artifacts. Indeed, in the final column of Fig. $\ref{fig:NYU_vis}$, we can see how the direct residual predictions produced by our networks also contain sharp features which cancels out the non-smoothness in the sparse maps.

\begin{table}[t!]
\centering
\caption{\textbf{Timing for D$^3$ and other architectures.} Models are evaluated assuming 0.18\% sparsity and using 1 Maxwell Titan X. The D$^3$ network achieves the lowest RMSE compared to other well known efficient network architectures. A slim version of D$^3$ runs at a near real-time 16fps for VGA resolution.}
\begin{tabular}{llllll|llllll}
\hline\noalign{\smallskip}
Model  & $L$ & $k$ & RMSE  & FPS  & Forward  &Model &  $L$ & $k$ & RMSE  & FPS  & Forward   \\
&& &(m)&&Pass (s)&&& &(m)&&Pass (s)\\
\noalign{\smallskip}
\hline 
\noalign{\smallskip}

D$^3$ & 5 & 12 & \textbf{0.118} & 10 & 0.11 &SegNet \cite{badrinarayanan2017segnet}& - & - & 0.150 & 5 & 0.20 \\
D$^3$ & 3 & 8 & 0.127 & 13 & 0.08 &ENet \cite{paszke2016enet}& - & - & 0.237 & 25 & 0.04\\
D$^3$ & 2 & 6 & 0.131 & 16 & 0.06 \\
\hline
\label{table:speed_comparison2}

\end{tabular}
\end{table}

\subsection{Computational Analysis}

In Table \ref{table:speed_comparison2} we show the forward pass time and accuracy for a variety of models at 0.18\% points sampled. Our standard D$^3$ model with $L=5$ and $k=12$ achieves the lowest error and takes 0.11s per VGA frame per forward pass. Slimmer versions of the D$^3$ network incur mild accuracy degradation but still outperform other well known efficient architectures \cite{badrinarayanan2017segnet,paszke2016enet}. The baseline speed for our D$^3$ networks can thus approach real-time speeds for full-resolution 480$\times$640 inputs, and we expect these speeds can be further improved by weight quantization and other optimization methods for deep networks  \cite{han2015deep}. Trivially, operating at half resolution would result in our slimmer D$^3$ networks operating at a real-time speed of $>$60fps. This speed is important for many application areas where depth is a critical component for scene understanding.

\subsection{Generalizing D$^3$ to Multiple Patterns and the Effect of $\mathcal{S}_2$}\label{sec:multires}

\begin{figure}[htb!]
\centering
	\begin{subfigure}[h]{0.45\linewidth}
		\includegraphics[width=\textwidth]{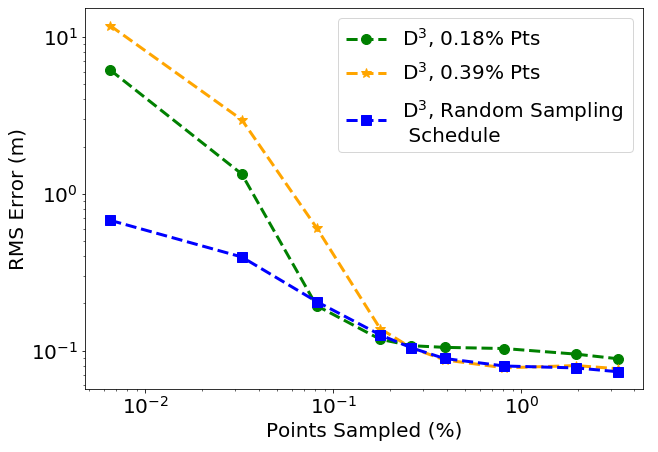} 
        \caption{}
	\end{subfigure}
	~
	\begin{subfigure}[h]{0.45\linewidth}
		\includegraphics[width=\textwidth]{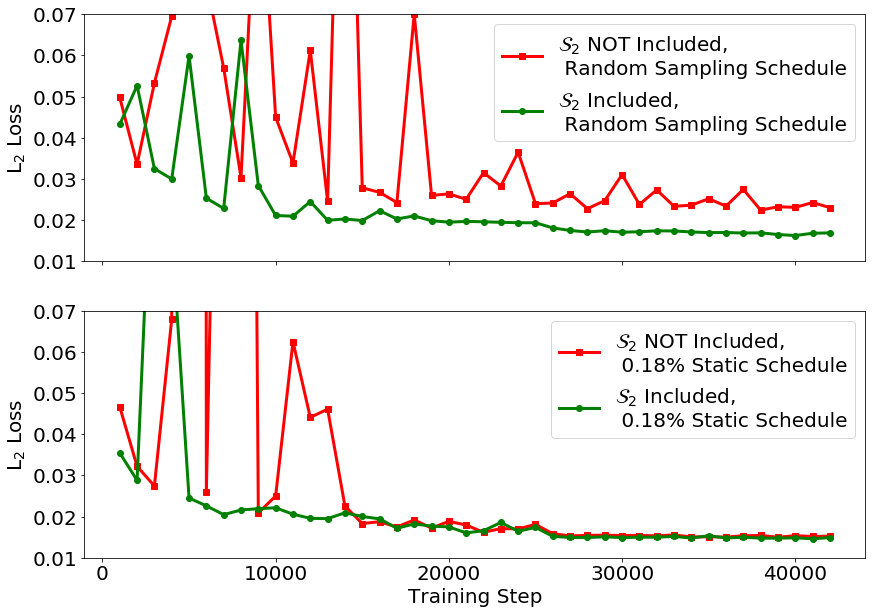}
        \caption{}
	\end{subfigure}
	\caption{\textbf{Multi-sparsity D$^3$ models.} (a) The Random Sampling network was trained with a different sparse pattern (between 0.065\% and 0.98\% points sampled) on every iteration, and performs well at all sparsity levels while being only mildly surpassed by single-density networks at their specialized sparsities. (b) Validation loss curves (for 0.18\% points sampled) for a D$^3$ models trained with and without inclusion of distance map $\mathcal{S}_2$. $\mathcal{S}_2$ is clearly crucial for stability and performance, especially when training with complex pattern schedules.}
	\label{fig:random_sampling}
\end{figure}

We train a D$^3$ network with a different input sparsity (sampled uniformly between 0.065\% and 0.98\% points) for each batch. Fig. \ref{fig:random_sampling}(a) shows how this multi-sparsity D$^3$ network performs relative to the 0.18\% and 0.39\% sparsity models. The single-sparsity trained D$^3$ networks predictably perform the best near the sparsity they were tuned for. However, the multi-sparsity D$^3$ network only performs slightly worse at those sparsities, and dramatically outperforms the single-sparsity networks away from their training sparsity value. Evidently, a random sampling schedule effectively regularizes our model to work \textit{simultaneously at all sparsities within a wide range}. This robustness may be useful in scenarios where different measurement modes are used in the same device. 

Inclusion of the distance map $\mathcal{S}_2$ gives our network spatial information of the sparse pattern, which is especially important when the sparse pattern changes during training. Fig. \ref{fig:random_sampling}(b) shows validation L$_2$ loss curves for D$^3$ networks trained with and without $\mathcal{S}_2$. $\mathcal{S}_2$ improves relative L$_2$ validation loss by 34.4\% and greatly stabilizies training when the sparse pattern is varied randomly during training. For a slow decay sampling schedule (i.e. what is used for the majority of our D$^3$ networks), the improvement is 8.8\%, and even for a static sampling schedule (bottom of Fig. \ref{fig:random_sampling}(b)) there is a 2.8\% improvement. The inclusion of the distance map is thus clearly essential to train our model well.

\subsection{Generalizing D$^3$ to Outdoor Scenes}\label{sec:KITTI_results}

We extend our model to the challenging outdoor KITTI dataset \cite{Uhrig2017THREEDV}. All our KITTI D$^3$ models are initialized to a pre-trained NYUv2 model. We then train either with only KITTI data (KITTI-exclusive) or with a 50/50 mix of NYUv2 and KITTI data for each batch (mixed model). Since NYUv2 images have a max depth of 10m, \textit{depth values are scaled by 0.1} for the KITTI-exclusive model. For the mixed model we use a scene-agnostic scaling rule; we scale all images down to have a max depth of $\leq$10m, and invert this scaling at inference. Our state-of-the-art results are shown in Table \ref{table:KITTI_results}. Importantly, as for NYUv2, our mixed model only performs slightly worse than the KITTI-exclusive network. More results for additional sparsities are presented in the Supplementary Material.
\begin{table}[htb!]
\caption{\textbf{D$^3$ Model Performance on the KITTI Dataset.} Lower values of RMSE and MRE are better, while higher values of $\delta_i$ are better. For competing methods we show results at the closest sparsity. The performance of our models, including the mixed models, is superior by a large margin.}
\begin{center}
\label{table:KITTI_results}
\begin{tabular}{llllllll}
\hline\noalign{\smallskip}
Model   & \% Points & Downsample&RMSE  & MRE  & $\delta_1$ & $\delta_2$ & $\delta_3$ \\
& Sampled & Factor &(m)&(\%)&(\%)&(\%)&(\%)\\
\noalign{\smallskip}
\hline
\noalign{\smallskip}
NN Fill & 0.077& 36$\times$36 &   4.441&9.306&91.88&97.75&99.04\\
D$^3$ Mixed (Ours) & 0.077& 36$\times$36 &   1.906&3.14&98.62&99.65&99.88\\
D$^3$ (Ours) & 0.077& 36$\times$36 &   \textbf{1.600}&\textbf{2.50}&\textbf{99.12}&\textbf{99.76}&\textbf{99.91}\\
\noalign{\smallskip}
\hline
\noalign{\smallskip}
Ma et al. \cite{ma2017sparse} & 0.096 & $\sim$32$\times$32 & 3.851 & 8.3 & 91.9 & 97.0 & 98.6\\
NN Fill & 0.174& 24$\times$24 &  3.203&5.81&96.62&99.03&99.57 \\
D$^3$ Mixed (Ours) & 0.174& 24$\times$24 &  1.472&2.22&99.30&99.83&99.94 \\
D$^3$ (Ours) & 0.174& 24$\times$24 &  \textbf{1.387}&\textbf{2.09}&\textbf{99.40}&\textbf{99.85}&\textbf{99.95} \\
\noalign{\smallskip}
\hline
\noalign{\smallskip}
Ma et al. \cite{ma2017sparse} & 0.240 & $\sim$20$\times$20 & 3.378 & 7.3 & 93.5 & 97.6 & 98.9\\
NN Fill & 0.391& 16$\times$16 &   2.245&3.73&98.67&99.60&99.81 \\
D$^3$ Mixed (Ours) & 0.391& 16$\times$16 &   1.120&1.62&99.67&99.92&99.97 \\
D$^3$ (Ours) & 0.391& 16$\times$16 &   \textbf{1.008}&\textbf{1.42}&\textbf{99.76}&\textbf{99.94}&\textbf{99.98} \\
\hline
 \\
\end{tabular}
\end{center} 
\end{table}

\begin{figure}[htb!]

\centering
\includegraphics[scale=0.44]{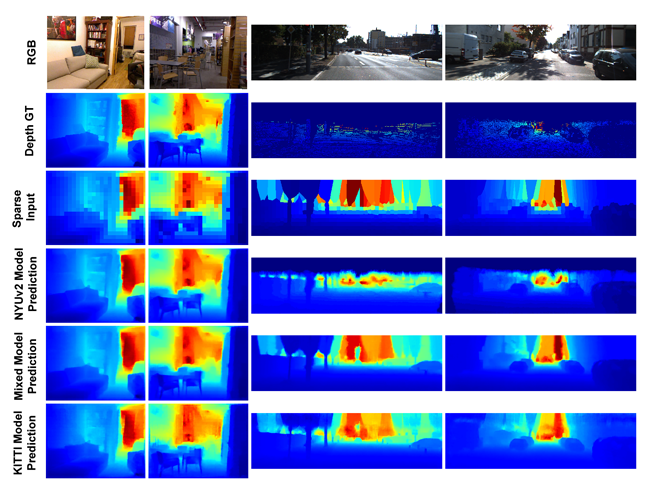}
\caption{\textbf{Joint Predictions on NYUv2 and KITTI.} The RGB, Depth GT, and Sparse Input $\mathcal{S}_1$ are given in the first three rows. Predictions by three models on both indoors and outdoors scenes are given in the final three rows, with the second-to-last row showing the mixed model trained on both datasets simultaneously. All sparse maps have a density of $0.18\%$ ($\sim$24$\times$24 downsampling).}
\label{fig:mixed_vis} 
\end{figure}
\newpage

Visualizations of the our model outputs are shown in Fig. \ref{fig:mixed_vis}. The highlight here is that the mixed model produces high-quality depth maps for both NYUv2 and KITTI.  Interestingly, even the KITTI-exclusive model (bottom row of Fig. \ref{fig:mixed_vis}) produces good qualitative results on the NYUv2 dataset. Perhaps more strikingly, even an NYUv2 pretrained model with no KITTI data training (third-to-last row of Fig. \ref{fig:mixed_vis}) produces reasonable results on KITTI. This suggests that our D$^3$ models intrinsically possess some level of cross-domain generalizability. 
\subsection{Robustness Tests}\label{sec:robustness}

Thus far, we have sampled depth from high-quality Kinect and LiDAR depth maps, but in practice sparse depth inputs may come from less reliable sources. We now demonstrate how our D$^3$ network performs given the following common errors within the sparse depth input:

\begin{enumerate}
\item Spatial misregistration between the RGB camera and depth sensor.
\item Random gaussian error.
\item Random holes (dropout), e.g. due to shadows, specular reflection, etc.
\end{enumerate}

\begin{figure}[htb!]
\centering
\includegraphics[scale=0.12]{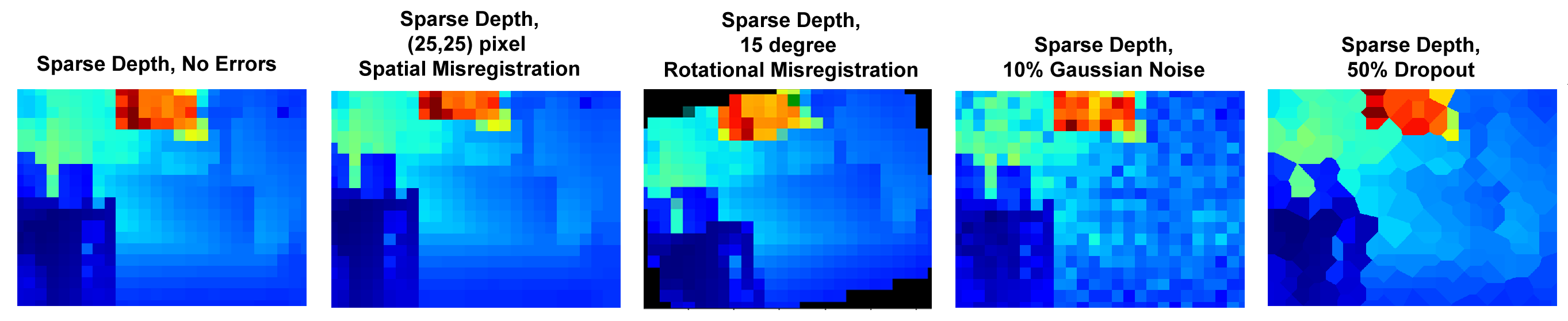}
\caption{\textbf{Potential Errors in Sparse Depth}. The three sparse depth maps on the right all exhibit significant errors that are common in real sensors.}
\label{fig:error_viz}
\end{figure}

\begin{figure}[htb!]
\centering
\includegraphics[scale=0.26]{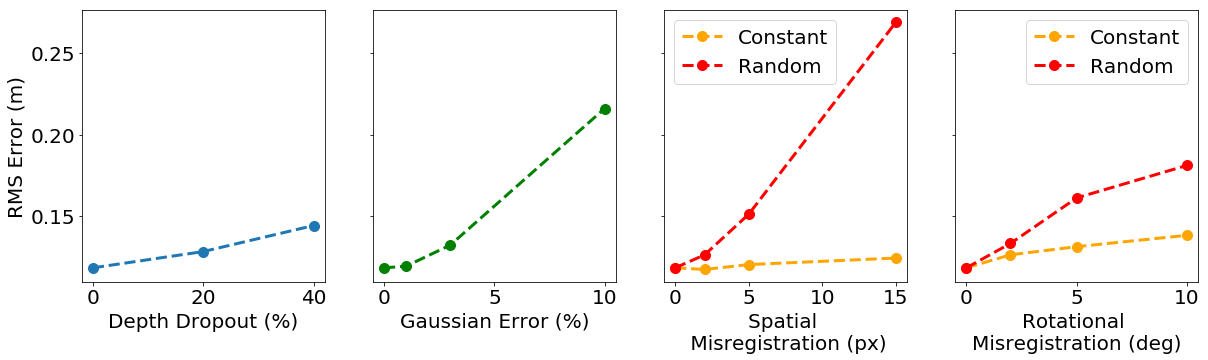}
\caption{\textbf{Accuracy of D$^3$ Networks under Various Sparse Depth Errors}. Under all potential error sources (with the exception of the unlikely random spatial mis-registration error), the D$^3$ network exhibits graceful error degradation. This error degradation is almost negligible for constant spatial mis-registration.}
\label{fig:error_plots}
\end{figure}

In Fig. \ref{fig:error_viz} we show examples of each of these potential sources of error, and in Fig. \ref{fig:error_plots} we show how D$^3$ performs when trained with such errors in the sparse depth input (see Supplementary Material for for tabulated metrics). The D$^3$ network degrades gracefully under all sources of error, with most models still outperforming the other baselines in Table \ref{table:NYU_results} (none of which were subject to input errors). It is especially encouraging that network performs robustly under constant mis-registration error, a very common issue when multiple imaging sensors are active in the same visual system. The network effectively learns to fix the calibration between the different visual inputs. Predictably, the error is much higher when the mis-registration is randomly varying per image.
 
\subsection{Discussion}
Through our experiments, we've shown how the D$^3$ model performs very well at taking a sparse depth measurement in a variety of settings and turning it into a dense depth map. Most notably, our model can simultaneously perform well both on indoor and outdoor scenes. We attribute the overall performance of the model to  a number of factors. As can be gathered from Table \ref{table:speed_comparison2}, the design of our multi-scale architecture, in which the sparse inputs are ingested at various scales and outputs are treated as residuals with respect to $\mathcal{S}_1$, is important for optimizing performance. Our proposed sparse input parameterization clearly allows for better and more stable training as seen in Fig. \ref{fig:random_sampling}. Finally, the design of the training curriculum, in which we use varying sparsities in the depth input during training, also plays an important role. Such a strategy makes the model robust to test time variations in sparsity (see Fig. \ref{fig:random_sampling}) and reduces overall errors. 


\section{Conclusions}

We have demonstrated that a trained deep depth densification (D$^3$) network can use sparse depth information and a registered RGB image to produce high-quality, dense depth maps. Our flexible parametrization of the sparse depth information leads to models that generalize readily to multiple scene types (working simultaneously on indoor and outdoor images, from depths of 1m to 80m) and to diverse sparse input patterns. Even at fairly aggressive sparsities for indoor scenes, we achieve a mean absolute relative error of under 1$\%$, comparable to the performance of consumer-grade depth sensor hardware. We also found that our model is fairly robust to various input errors.

We have thus shown that sparse depth measurements can be sufficient for applications that require an RGBD input, whether indoors or outdoors. A natural next step in our line of inquiry would be to evaluate how densified depth maps perform in 3d-reconstruction algorithms, tracking systems, or perception models for related vision tasks such as surface normal prediction. We hope that our work motivates additional research into uses for sparse depth from both the software and hardware perspectives.


\clearpage


\newpage
\section{Estimating Depth from RGB and Sparse Sensing: Supplementary Materials}
\subsection{Averaging versus Point Sampling}

We chose (as in \cite{lu2015sparse}) in our problem setup to sample our sparse depths from the GT depth pointwise, as opposed to averaging sparse depths in a patch around each sampled point. We do this because sparsifying depth by a factor of A$\times$A should imply that we save a factor of $A^2$ in power for data collection, which indicates that sparse sensor should not have access to any other depth information other than the sampling points. 

\begin{table}[htb!]
\centering
\caption{D$^3$ Model Performance for Averaging}
\label{table:averaging_results}
\begin{tabular}{llllll}
\hline\noalign{\smallskip}
Patch   & RMSE  & MRE  & $\delta_1$ & $\delta_2$ & $\delta_3$ \\
Size (px)& (m)&(\%)&(\%)&(\%)&(\%)\\
\noalign{\smallskip}
\hline
\noalign{\smallskip}
1 & 0.118 & 1.49 & 99.45 & 99.92 & 99.98 \\
3 & 0.117 & 1.46 & 99.46 & 99.92 & 99.99 \\
5 & 0.111 & 1.42 & 99.53 & 99.94 & 99.99 \\
11 & 0.102 & 1.35 & 99.63 & 99.95 & 99.99 \\
\hline
\end{tabular}
\end{table}

However, we do present results for the averaging scenario in Table \ref{table:averaging_results}: our sparse input $\mathcal{S}_1(x,y)$ at a sample point $(x,y)$ is not a point sample $D(x,y)$, but rather an average of $D$ around a point $(x,y)$. Predictably, by adding in this additional information we steadily decrease our network error, although this comes at a cost of having to exert much more effort in collecting the sparse input; for a relative improvement of $\sim 15\%$ in RMSE we must gather 120 times more data corresponding to an 11$\times$11 patch around each center point. 

\subsection{Additional Results on KITTI}
\begin{figure}[htb!]
 	\begin{subfigure}[h]{0.45\linewidth}
 		\includegraphics[width=\textwidth]{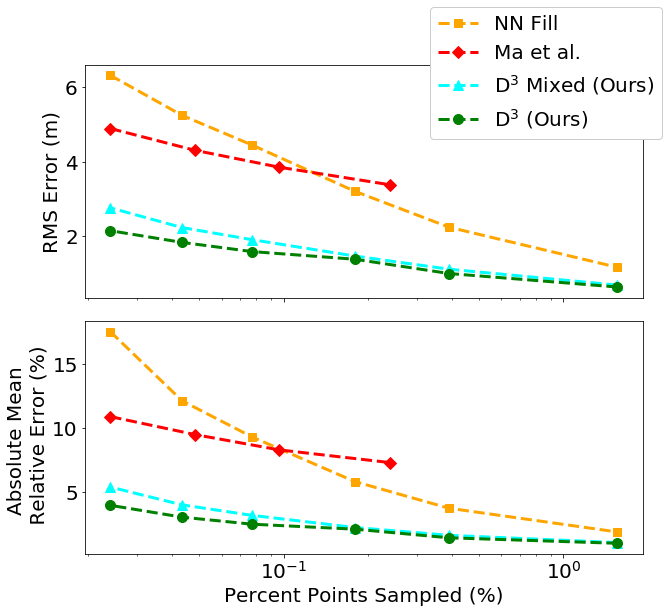}       						   
 	\end{subfigure}
 	~
 	\begin{subfigure}[h]{0.45\linewidth}
 		\includegraphics[width=\textwidth]{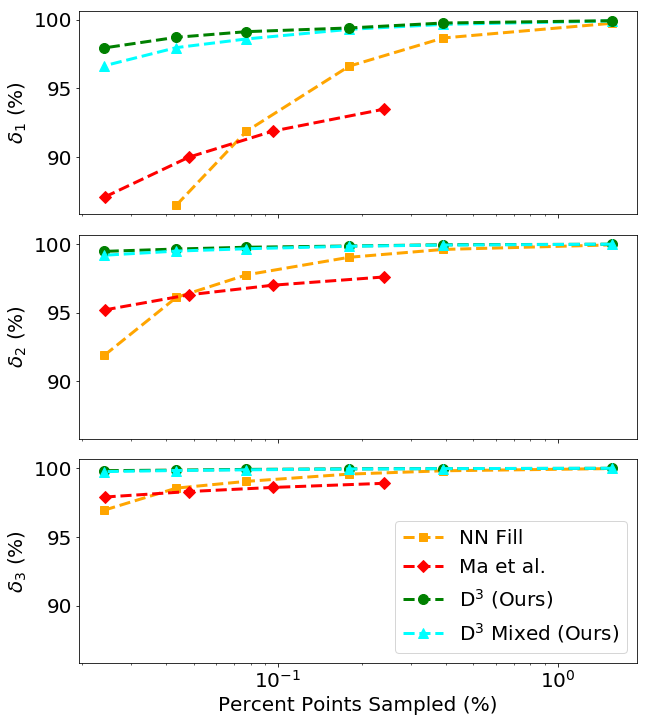}
 	\end{subfigure}
 	\caption{\textbf{Results on the KITTI Dataset.} Plots are in the same layout as in Fig. \ref{fig:nyu_plots} in the main paper, with RMSE and MRE plotted on the left and the $\delta_i$ are plotted on the right. Lower values of RMSE and MRE are better, while higher values of the $\delta_i$ are better. Our D$^3$ network outperforms baselines and other methods, and dramatically so at lower pattern densities.}
 	\label{fig:KITTI_graphs}
 \end{figure}

Here we provide additional results on the KITTI dataset for more sparsity levels (Fig. \ref{fig:KITTI_graphs}). We see that our D$^3$ models substantially outperform all other methods/baselines.
 \pagebreak
\subsection{Additional Results for Tighter $\delta$ Thresholds}

\begin{figure}[htb!]
\centering
	\begin{subfigure}[h]{0.45\linewidth}
		\includegraphics[width=\textwidth]{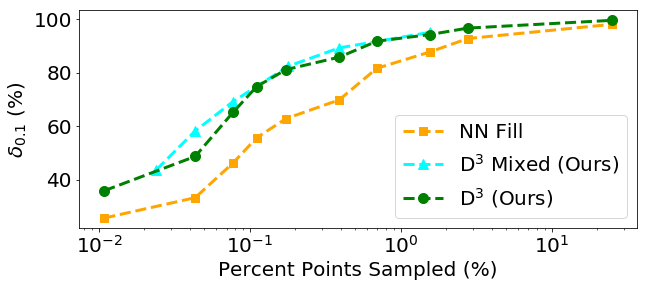} 
        \caption{}
	\end{subfigure}
	~
	\begin{subfigure}[h]{0.45\linewidth}
		\includegraphics[width=\textwidth]{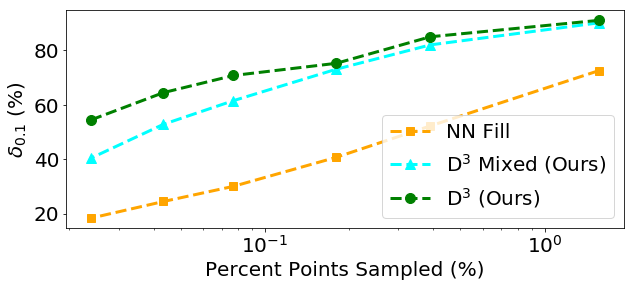}
        \caption{}
	\end{subfigure}
	\caption{\textbf{Additional $\delta$ metrics for tighter $\delta$ thresholds.} (a) Tighter $\delta_{0.1}$ results on NYUv2. (b) Tighter $\delta_{0.1}$ results on KITTI}
	\label{fig:tighter_delta}
\end{figure}

Although a standardly reported metric for depth estimation is $\delta_i$ for $i\in \{1,2,3\}$, we found that such $\delta$ bounds are too tight to see a meaningful curve on our $D^3$ results, as most $\delta_i$ values for $D^3$ lie too close to $100\%$. Although it's rare in the literature to report $\delta_i$ bounds for $i<1$, we present in Figure \ref{fig:tighter_delta} some results on $\delta_{0.1}$, which corresponds to a threshold value of approximately $\approx 1.02$. We see that the results are fairly consistent with the results presented in the main paper: $D^3$ significantly outperforms the sparse NN fill baseline, and the mixed and exclusive models both perform similarly well. 

\subsection{Additional Ablation Studies on $D^3$ Design Choices}\label{sec:additional_ablation}
In this section we present additional ablation studies on some of the more meaningful design choices we picked when training our $D^3$ network. Namely, there are two choices which we investigate: 
\begin{enumerate}
\item Including the sparse depth map $\mathcal{S}_1$ at every DenseNet module within the $D^3$ architecture (as opposed to only at the input layer). 
\item Performing a nearest-neighbor fill on $\mathcal{S}_1$ (as opposed to inputting the raw depth maps). 
\end{enumerate}
\begin{table}[htb!]
\centering
\caption{D$^3$ Additional Ablation Studies. All experiments run at 0.18\% sparsity. The ablation numbering refers to the numbering in the ablation enumeration in Section \ref{sec:additional_ablation}. Standard sparsity schedule refers to the sparsity schedule used in the main paper (in Section \ref{sec:sampling_strategies}). Random sparsity schedule refers to randomly choosing a sparsity at each training batch.	}
\label{table:ablation}
\begin{tabular}{l|l|l}
\hline\noalign{\smallskip}
\textbf{Sparsity Schedule}&\textbf{Ablation}&\textbf{RMSE (m)}\\
\noalign{\smallskip}
\hline
\noalign{\smallskip}
Standard&None &\textbf{0.118} \\
Standard&1& 0.120\\
Standard&2& 0.287\\
\hline 
Random&None &\textbf{0.125}\\
Random&1&0.135\\
Random&2&0.344\\
\hline
\end{tabular}
\end{table}

We tabulate the results of the ablation studies in Table \ref{table:ablation}. We see that all our design choices were appropriate in that ablating said choices degraded model performance in at least one important setting. When we do not include the sparse depth map $\mathcal{S}_1$ at every DenseNet layer, the performance is essentially the same when the training schedule (in terms of sparsity patterns) is fixed, but we see a significant degradation of performance when the training schedule is randomized. We also see a substantial degradation in model performance under ablation 2, in which we feed in raw sparse maps without any nearest-neighbor infilling. Interestingly the performance of the model under this ablation is closer to the performance of the methods suggested in \cite{ma2017sparse}, which demonstrates that our particular input parametrization is a powerful one.

\subsection{Tabulated Accuracies for Robustness Tests}

For further reference, we present here specific performance metric values for the robustness tests performed in Section \ref{sec:robustness}. 

\begin{table}[htb!]
\begin{center}
\caption{D$^3$ Model Performance for Various Errors in Sparse Input}
\label{table:error_results}
\begin{tabular}{lllllll}
\hline\noalign{\smallskip}
Error Type & Error & RMSE  & MRE  & $\delta_1$ & $\delta_2$ & $\delta_3$ \\
& Magnitude &(m)&(\%)&(\%)&(\%)&(\%)\\
\noalign{\smallskip}
\hline
\noalign{\smallskip}
No Error & 0 & 0.118 & 1.49 & 99.45 & 99.92 & 99.98 \\
\hline
Depth Dropout  &  20\%& 0.128 &1.69&99.33 & 99.90 & 99.98\\
& 40\% & 0.144 & 2.04 & 99.13 & 99.86 & 99.97\\
\hline
Gaussian Error & 1\% &0.119 &1.66&99.45&99.92&99.99\\
& 3\% &0.132&2.19&99.43&99.92&99.99\\
& 10\% &0.216&3.86&98.89&99.90&99.98 \\
\hline
Spatial Misregistration  & 2 px & 0.117&1.46&99.46&99.92&99.98\\
(Constant) & 5 px&0.120&1.53&99.44&99.92&99.98\\
& 15 px &0.124&1.70&99.42&99.92&99.98 \\
\hline
Spatial Misregistration  &  2 px & 0.126&1.67&99.35&99.90&99.98
\\

(Random) & 5 px&0.151&2.19&99.04&99.83&99.97\\
& 15 px &0.269&4.59&97.04&99.32&99.80 \\
\hline
Rotational Misregistration  & 2 deg & 0.126&1.62&99.38&99.91&99.98\\
(Constant) & 5 deg&0.131&1.89&99.30&99.90&99.98\\
& 10 deg &0.137&2.24&99.18&99.89&99.98 \\
\hline
Rotational Misregistration  & 2 deg & 0.132&1.82&99.29&99.89&99.98\\
(Random) & 5 deg&0.161&2.42&98.92&99.82&99.96\\
& 10 deg &0.184&3.11&98.52&99.74&99.94 \\
\hline
\end{tabular}
\end{center}
\end{table}

\newpage
\subsection{Preliminary Studies on Interest Point Sampling Maps, and Discussion of Some $D^3$ Weaknesses}

\begin{figure}[htb!]
\centering
\includegraphics[scale=0.5]{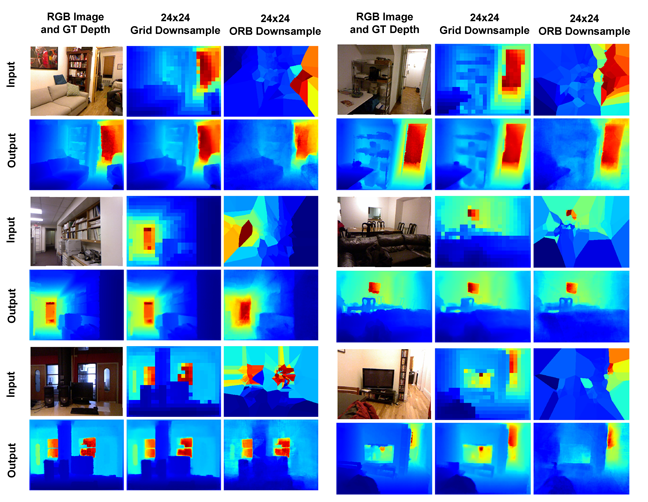}
\caption{Visualizations for ORB Interest Point Sparse Inputs.}
\label{fig:orb_vis}
\end{figure}

Our proposed flexible sparse input parametrization allows us to easily explore different sparse patterns for the sparse input depth. One potentially useful avenue of exploration is to explore standard interest point patterns such as ORB, as these sparse depth maps may be accessible through traditional SLAM systems. We present some results on ORB sampling in Table \ref{table:ORB_results} and Fig. \ref{fig:orb_vis}. 

\begin{table}[htb!]
\caption{\textbf{D$^3$ Model Performance on the NYUv2 Dataset for ORB Sampling}. For RMSE and MRE, lower values are better. For $\delta_i$ higher values are better.}
\begin{center}
\label{table:ORB_results}
\begin{tabular}{llllllll}
\hline\noalign{\smallskip}
Model   & \% Points & Downsample&RMSE  & MRE  & $\delta_1$ & $\delta_2$ & $\delta_3$ \\
& Sampled & Factor &(m)&(\%)&(\%)&(\%)&(\%)\\
\noalign{\smallskip}
\hline
\noalign{\smallskip}
Eigen et al. \cite{eigen2015predicting} & 0& N/A &  0.641&15.8&76.9&95.0&98.8 \\
Laina et al. \cite{laina2016deeper} & 0& N/A & 0.573 & 12.7 & 81.1& 95.3 & 98.8 \\
D$^3$ No Sparse & 0& N/A &  0.711&22.37&67.32&89.68&96.73 \\
\noalign{\smallskip}
\hline
\noalign{\smallskip}
NN Fill ORB & 0.174& 24$\times$24 & 0.749 & 17.20 & 78.25 & 91.19 & 96.34 \\
D$^3$ ORB & 0.174& 24$\times$24 & 0.347 & 8.90 & 91.69 & 97.86 & 99.35 \\

D$^3$ GRID & 0.174& 24$\times$24 & 0.118 & 1.49 & 99.45 & 99.92 & 99.98 \\
\hline
\end{tabular}
\end{center}
\end{table}

ORB sampling performs worse than regular grid sampling, and the visualizations show some edge artifacting in the depth maps produced from the highly non-smooth sparse inputs (however, considering the high variability of these sparse inputs, it is actually quite impressive that the outputs have normalized out these artifacts as much as they have). This performance gap is natural, considering interest point sampling has significantly lower coverage compared with regular grid sampling (and even compared with random sampling). Interest point sampling also tends to focus on corner or other difficult features in the image; having sparse depth available at these difficult points with significant depth variation is less helpful. However, interest point sampling has desirable consistency properties that can reduce temporal jitter in the depths maps produced, thus providing an interesting avenue of future study.

In general, as the sparsity maps become highly irregular, our $D^3$ network tends to struggle to fully smooth out its final depth predictions. Highly irregular patterns also often result in sparse maps that entirely miss object instances within a particular image, and we've found that in this case it is difficult to recover that object instance with a $D^3$ forward pass. These are two known weaknesses of our work thus far, and we are experimenting with additional smoothness losses and inputting more meaningful semantic information (in the hopes of detecting when object instances do not intersect with our sparse maps) into our $D^3$ framework to mitigate both these effects.

\end{document}